\documentclass[runningheads]{llncs}

\usepackage[mobile]{eccv}

\usepackage{eccvabbrv}

\usepackage{graphicx}
\usepackage{booktabs}

\usepackage[accsupp]{axessibility}  

\usepackage{hyperref}

\usepackage{orcidlink}

\usepackage{overpic}
\usepackage{booktabs} 
\usepackage{subcaption,graphicx}

\usepackage{comment}
\usepackage{booktabs}
\usepackage{multirow}
\usepackage{mathtools}
\usepackage{cancel}
\usepackage{multirow}
\usepackage{amsmath}
\usepackage{amssymb}
\usepackage[separate-uncertainty = true,multi-part-units=single]{siunitx}
\usepackage{algorithm,algpseudocode}

\usepackage{float}  
\usepackage{stfloats}  
\usepackage{wrapfig}   

\usepackage{color}
\definecolor{blue}{rgb}{0,0,0.6}
\definecolor{green}{rgb}{0,0.6,0}
\definecolor{red}{rgb}{0.6,0,0}
\definecolor{gray}{rgb}{0.4,0.4,0.4}
\definecolor{black}{rgb}{0,0,0}
\definecolor{lightgray}{rgb}{0.83, 0.83, 0.83}
\definecolor{purple}{rgb}{1,0,1}

\usepackage{enumitem}

\renewcommand{\vec}[1]{\mathbf{#1}}
\newcommand{\mat}[1]{\mathbf{#1}}

\usepackage{dsfont}
\newcommand{\R}{\mathds{R}}

\newcommand{\pcamat}{\mat{U}}
\newcommand{\pcamean}{\bar{\vec{u}}}
\newcommand{\pcaparams}{\vec{w}}

\begin{document}

\title{GarmentCodeData: A Dataset of 3D Made-to-Measure Garments With Sewing Patterns} 

\titlerunning{GarmentCodeData}

\author{
Maria Korosteleva\inst{1}\orcidlink{0000-0001-7151-0946} \and 
Timur Levent Kesdogan\inst{1}\orcidlink{0009-0006-5839-9677} \and   %
Fabian Kemper\inst{2}\orcidlink{0009-0002-0303-417X}  \and 
Stephan Wenninger\inst{2}\orcidlink{0009-0008-2404-7117}  \and
Jasmin Koller\inst{1}\orcidlink{0009-0004-7307-3566}  \and 
Yuhan Zhang\inst{1}\orcidlink{0009-0001-0686-3865}  \and 
Mario Botsch\inst{2}\orcidlink{0000-0001-9954-120X} \and \\
{Olga Sorkine-Hornung}\inst{1}\orcidlink{0000-0002-8089-3974}    
}

\authorrunning{M.~Korosteleva et al.}

\institute{ETH Zurich, Rämistrasse~101, 8092 Zurich, Switzerland 
\email{\{maria.korosteleva,olga.sorkine\}@inf.ethz.ch} \\
\email{\{tkesdogan,kollerja,yuhzhang\}@student.ethz.ch}   \and
TU Dortmund University, Otto-Hahn-Strasse 16, 44227 Dortmund, Germany
\email{\{fabian.kemper,stephan.wenninger,mario.botsch\}@tu-dortmund.de}\\
}

\maketitle

\begin{abstract}
    Recent research interest in learning-based processing of garments, from virtual fitting to generation and reconstruction, stumbles on a scarcity of high-quality public data in the domain. We contribute to resolving this need by presenting the first large-scale synthetic dataset of 3D made-to-measure garments with sewing patterns, as well as its generation pipeline. GarmentCodeData contains 115,000 data points that cover a variety of designs in many common garment categories: tops, shirts, dresses, jumpsuits, skirts, pants, etc., fitted to a variety of body shapes sampled from a custom statistical body model based on CAESAR~\cite{robinette2002caesar}, as well as a standard reference body shape, applying three different textile materials. 
    To enable the creation of datasets of such complexity, we introduce a set of algorithms for automatically taking tailor's measures on sampled body shapes, sampling strategies for sewing pattern design, and propose an automatic, open-source 3D garment draping pipeline based on a fast XPBD simulator~\cite{warp2022}, while contributing several solutions for collision resolution and drape correctness to enable scalability. 
  
\end{abstract}

\section{Introduction}

Reconstructing CAD representation of garments -- sewing patterns -- is an increasingly relevant problem. The sewing pattern is a key garment representation that not only connects design with production, but is also essential to the fitting process, be it physical or virtual, allowing detailed garment editing and application of various appearance models and fabrics. The advancements of generative models, virtual fitting rooms and reconstruction of photorealistic full-body avatars all require scalable solutions for obtaining garment rest shapes from raw inputs -- images and 3D models. Learning-based solutions have started to appear \cite{Korosteleva2022,Liu2023TowardsImage}, but the progress is hindered by the lack of open, large-scale datasets of 3D garments labeled with sewing patterns. 


\begin{figure}
  \centering
  \begin{overpic}[trim=0cm 0cm 0cm 0cm,clip,width=1.\linewidth,grid=false]{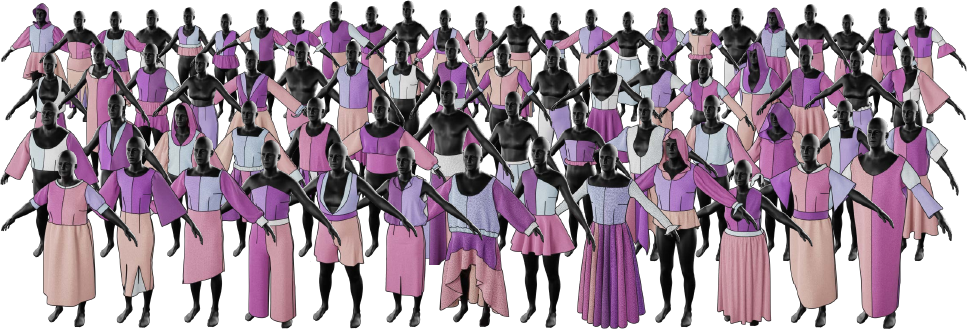}  %
    \end{overpic}
  \caption{GarmentCodeData samples covering a variety of designs and body shapes.} 
  \label{fig:teaser}
\end{figure}

Gathering datasets of the required scale and diversity from real-world data faces major challenges w.r.t.\ copyright, paywalls and general unavailability of garment-pattern pairs. Synthetic datasets~\cite{Bertiche2020,Korosteleva2021,Liu2023TowardsImage} provide an alternative that does not face these issues, while offering high-quality data and predictable coverage of the domain. However, generating a reasonably representative set is still a difficult problem. In this work we address the problem of design and body shape variation in synthetic garment generation, creating the first large-scale open dataset of made-to-measure garments with our generation pipeline.  

Our design sampling process is based on the parametric modeling approach of GarmentCode~\cite{Korosteleva2023GarmentCode:Patterns}, with the addition of sampling probabilities to the design variation parameters and the editing of the design space itself to reduce the chances of randomly creating invalid samples (e.g., skirts twice as long as the height of a person). Through consultations with a professional fashion designer, we have also improved the quality of the sewing patterns originally implemented in GarmentCode and expanded the set to include new elements, e.g.\ hoods. As for body shape variation, we utilize a statistical body model and introduce a set of algorithms for automatically evaluating body measurements from body shape samples. The dataset is obtained by random sampling of body shapes and design parameters to generate 
sewing patterns, which are then draped on the bodies by our automatic draping pipeline, based on the extended position-based dynamics (XPBD)  simulator~\cite{Macklin2016XPBD:Dynamics,warp2022}. 
To achieve the required quality of drapes across body shapes and designs, we introduce several solutions to combat the issues that arise, including cloth-cloth and body-cloth constraints that help to resolve collisions in the initial sewing pattern placement. For each generation step, we implement quality control conditions to exclude bad samples from the data.

The final dataset consists of 115,000 design samples, each fitted to a random body shape and a canonical neutral body model (gender-neutral statistical average), labeled with a ground truth sewing pattern and a 3D draped mesh, alongside segmentation labels, a UV map corresponding to initial panel shapes, body measurements, design parameters and basic renders for visualization. Designs span a wide variety of lower-body (skirts and pants of different styles), upper-body (tops, shirts, blouses), and full-body (dresses, jumpsuits) garments. 

Our work presents the following contributions: 
\vspace{-2mm}
\begin{itemize}
    \item Controllable parametric design space of quality sewing patterns, suitable for garment design sampling. 
    \item Fit-aware automatic body measurement method.
    \item Open-source pipeline for generating 3D garment models from sewing patterns at scale.
    \item The first open, large-scale dataset of made-to-measure garments. It covers a variety of complex designs and body shapes, and provides 3D models and corresponding sewing patterns with each data point.
\end{itemize}

The links to the dataset and available code are shared on the \href{https://igl.ethz.ch/projects/GarmentCodeData/}{project page.}\hspace{-.12em}\footnote{\url{https://igl.ethz.ch/projects/GarmentCodeData/}}

\section{Related work}

\vspace{-2mm}
\noindent\textbf{3D garment datasets.\ }
Existing datasets of 3D garments or clothed humans can be put into three categories: those based on 3D scans of physical garments~\cite{Zhang2017a,LalBhatnagar2019,Ma2020,Zhu2020,Tiwari2020,Antic2024CloSe:Model,Ho2023LearningHumans,Xu2023ClothPose:Solution}, synthetic~\cite{Wang2011Data-DrivenMeasurement,Narain2012AdaptiveSimulation,Wang2018,Pumarola2019a,Jiang2020,Bertiche2020,Vidaurre2020,Korosteleva2021,Liu2023TowardsImage}, and datasets with garments manually created by professional designers~\cite{Black2023BEDLAM:Motion,Zou2023CLOTH4D:Reconstruction}. Datasets of real garments capture their realistic behavior, but are expensive to collect, often contain noise and face copyright issues when sharing, e.g.~\cite{Antic2024CloSe:Model}, which contains samples from commercial repositories. The design variation they provide is also limited, often focusing on common clothing pieces -- t-shirts, jeans, hoodies -- not capturing the diversity of designs available in the wild. 
To this day, Deep Fashion3D~\cite{Zhu2020} is the most diverse dataset of captured garments, but it contains only 563 items. 
These datasets are \emph{unusable} for sewing pattern reconstruction since they do not provide the sewing patterns corresponding to the 3D garments. 

The garments created in virtual CAD by professional designers in~\cite{Black2023BEDLAM:Motion,Zou2023CLOTH4D:Reconstruction} represent clothing of high complexity and allow seamless integration of these garments into generation pipelines for synthetic images. Unfortunately, this approach is hard to scale due to the labor resources for each piece. This is also a reason why a diverse set of body shapes is hard to achieve in this approach -- due to the need to manually fit a garment to a desired body shape. Garment sewing patterns should be available for these datasets in theory, but de facto they are shared in proprietary file formats that cannot be processed outside of the corresponding CAD systems. 

In this work, we opt for a synthetic dataset. This is a scalable solution with fully controlled output data distribution and high-quality data samples, well-aligned across the dataset, and with exact ground truth labels. Several synthetic garment datasets exist, but most do not supply sewing pattern labels at all~\cite{Pumarola2019a,Jiang2020,Bertiche2020,Zhou2023ClothesNet:Environment}, or cover only a small selection of garment styles~\cite{Wang2011Data-DrivenMeasurement,Narain2012AdaptiveSimulation,Wang2018,Vidaurre2020}. Korosteleva and Lee~\cite{Korosteleva2021} and Liu et al.~\cite{Liu2023TowardsImage} are the closest works to our new dataset that provide sewing patterns across a wide selection of garment designs, with the latter~\cite{Liu2023TowardsImage} using the former~\cite{Korosteleva2021} as the backbone. Korosteleva and Lee~\cite{Korosteleva2021}  laid significant groundwork for sewing pattern-based generation, but their dataset lacks in complexity of generated garments and does not provide variation of body shapes. Their generation pipeline is based on a commercial CPU-based physics simulator, limiting its adoption. Inspired by this work, we aim to address its issues. We introduce made-to-measure design samples fitted to a variety of body shapes, increase the complexity of garment designs, add small variations to poses for robustness, and implement the generation pipeline using an open-source GPU-based simulator~\cite{warp2022}, allowing for easier adaptation and scalability. We provide a significantly larger dataset consisting of 115,000 designs compared to 23,500 in Korosteleva and Lee~\cite{Korosteleva2021}.\\

\vspace{-2mm}
\noindent\textbf{Automatic body measurements.\ }
To accurately generate clothing for a specific body shape, the fitting of the sewing patterns should be informed by anthropometric measurements on the body.
Various methods for body shape measurement have been proposed:
Pujades et al.~\cite{Pujades2019TheMeasurements} use a VR tracking system and ask the user to place a VR controller on various locations on their body, thereby only requiring minimal data input. 
Given a 3D scan of a person, body measurements can be directly inferred from the corresponding 3D point cloud \cite{Lu2008AutomatedScanners}. 
Registering a statistical template model to 3D scans allows to define landmarks and piecewise-linear curves (polylines) on the template topology, which can be used to guide the measurement algorithms \cite{Bartol2022LinearMeasurement, Li2022RemodelingParameters, NourbakhshKaashki2021Anet:Extraction}.
From these template-based methods, models for regressing anthropometric measurements directly from (silhouette) images can be trained \cite{Ruiz2022HumanAugmentation3DV}.
Typically, axis-aligned planes and Euclidean as well as geodesic distances between landmarks are used to provide a set of measurements. 
Our body shape measurements are taken in a similar fashion, but we additionally optimize the predefined polylines to adjust them to the given body. 

\section{Data Generation Pipeline}

\subsection{Overview}

The data generation process consists of three steps: body shape sampling, sewing pattern sampling, and draping. We sample 5,000 body shapes from a custom statistical body model, and then automatically evaluate the body measurements for each sample with a collection of algorithms designed for different types of body measurements (\cref{sec:gen:body}). Apart from body shape variations, we include small variations in the arm angles to increase the robustness of the models trained on our dataset to slight variations in A-poses in the inputs. Next, sewing pattern designs are sampled (\cref{sec:gen:design}). We carefully adjust the design space to ensure a diverse yet reasonable design distribution in the final dataset. Every design sample is fitted to the canonical neutral body model (statistical average across genders) and to one of the randomly sampled body shapes based on the evaluated measurements, which provides a ground truth design reference in canonical shape for every sample. The generated sewing patterns are then draped on top of their corresponding body models using our custom draping pipeline, implemented using Warp~\cite{warp2022}, a GPU-based XPBD simulator. 

\subsection{Design space preparation and sampling}
\label{sec:gen:design}

\begin{wrapfigure}{r}{0.4\linewidth}%
  \includegraphics[width=\linewidth]{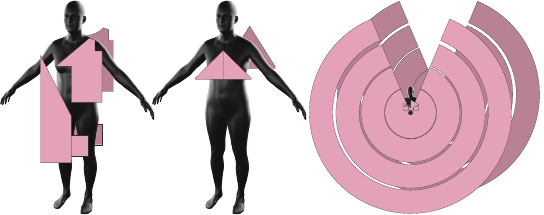}%
  \vspace{-0.2cm}
  \caption{Samples without guidance.}
  \label{fig:err_design_samples}%
\end{wrapfigure}

GarmentCode~\cite{Korosteleva2023GarmentCode:Patterns} provides a rich design space of garments. However, a naive uniform sampling of its design parameters results in numerous non-representative design samples, e.g., skirts or sleeve elements longer than the body dimensions, or unsuitable garment appearance, see~\cref{fig:err_design_samples}. We update the design space to ensure the sampling of interesting but reasonable garments.\\

\setlength{\columnsep}{10pt}%
\setlength{\intextsep}{0pt}%

\vspace{-2mm}
\noindent \textbf{Aligning the design parameters.\ }
First, we align the design variations with body measurements and with each other, e.g., lower body elements' lengths are kept within the leg length, and cuff elements' lengths are kept smaller than the length of corresponding sleeve or pant elements.\\

\vspace{-2mm}
\noindent \textbf{Sampling probability.\ } 
Secondly, we notice that usually, most of the design elements in daily wear follow a common style convention, with only some elements varying in style to add uniqueness to a particular piece. We implement this idea by introducing sampling probabilities to the default value of each design parameter $P_i$. We flip a coin with a probability $p_i$ to draw the default value of $P_i$. If the default value is not selected, we draw a non-default value from the corresponding parameter range uniformly.
The default values and their probabilities are defined manually to ensure reasonable and interesting designs and more common features to appear more often in the data samples. By adjusting the default value probabilities for the top-level garment structure, we set up roughly a third of the samples to be upper-body garments, another third -- lower-body garments, and the rest to be full-body garments (dresses and jumpsuits). We define a strong preference for symmetric garments by setting $p_i=0.8$ for the parameter that controls left-right asymmetry in necklines and sleeves.\\

\vspace{-2mm}
\noindent \textbf{Updating the template patterns.\ }
Finally, with the help of a professional fashion designer, we improve the default sewing patterns of the fitted garment elements (bodice, pencil skirt, pants) provided with GarmentCode in terms of the fit quality and pattern-making conventions alignment. We also add a hood element to represent hoodies in our dataset. 

Another important modeling consideration is the initial placement of the pattern panels (representing pieces of fabric) around the body, which heavily influences simulation quality (see~\cref{sec:eval:sim_fails}). We start by placing the highest-located garment elements relative to body locations, e.g., the top of an upper-body garment is placed at the neck level, tops of waistbands or lower-body garments are placed at the waist level when upper-body elements are not present, sleeves are rotated to align with arm pose angle. The remaining pieces are placed based on stitching information in relation to already placed components in order to reduce the chances of inter-panel collisions, 
e.g., lower-body garment elements are placed below a waistband, if present, and cuff elements are placed below corresponding sleeve or pant components.

We also expand the functionality of GarmentCode by introducing the ability to add labels to panels or individual panel edges, which facilitates the simulation process by providing references for attachment constraints and body part segmentation, as described below.\\

\vspace{-2mm}
\noindent \textbf{Quality control.\ }
Incompatibility of some parameter combinations is a common pitfall of complex parametric modeling systems, ours included. After a design sample is drawn, we employ additional quality control procedures to remove inadequate samples that could not be prevented with the measures above. We exclude garment samples that contain self-intersecting panels, full-body garments whose total length exceeds the floor length (distance from the base of the neck to the bottom of the feet) for a given body shape, empty patterns, garments that only consist of a belt, and lower garments with a high chance of sliding off. The latter include skirts with the upper circumference wider than the body trunk, and heavy (wide or long) flare skirts with a very thin or absent belt. The ``heaviness'' is defined heuristically based on our experiments of simulating flare skirts with different design parameters.

\subsection{Automatic body measurements}
\label{sec:gen:body}

\noindent\textbf{Body shape sampling.}
We generate a statistical human shape model by registering a template mesh to the European subset of the CAESAR database~\cite{robinette2002caesar}, consisting of 1700 3D scans annotated with anthropometric measurements. 
To register our template to the database, we follow the common template fitting approach~\cite{loper2015SMPL, registration-tutorial, achenbach17}, yielding 1700 surface meshes with the same topology and pose. 
We apply principal component analysis (PCA) to the registered meshes to obtain a low-dimensional shape space, which we sample to generate 5000 body shapes (see details in the supplemental material).\\

\begin{figure}[t]
    \centering
    \includegraphics[width=1.\linewidth]{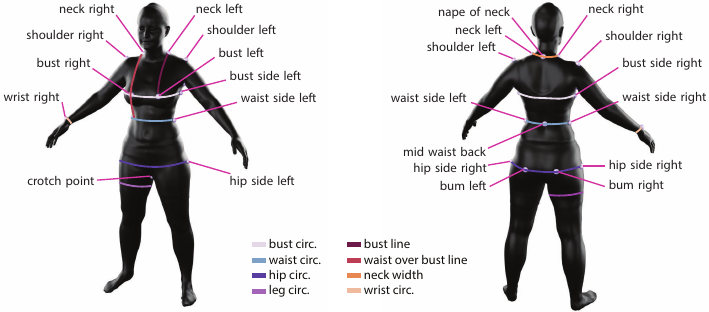}
    \caption{Overview of the body landmarks and measurements. } 
    \label{fig:landmarks}
\end{figure}

\vspace{-2mm}
\noindent\textbf{Body measurements.\ }
We automatically compute the body measurements required to generate garment patterns for the sampled body shapes. These serve as input data to fit the updated GarmentCode~\cite{Korosteleva2023GarmentCode:Patterns} patterns to the bodies. {One of the main rules in garment fit is balance: the main feature lines should look straight, e.g., the side stitch is vertical, and the waistline is parallel to the ground, making the wearer's posture look straight and upright. To support this goal we take the body measurements directly along the balance lines, ensuring correctly fitted sewing patterns as a result. Hence, most measurements are taken parallel to the ground or vertical direction.} To this end, we select predefined landmarks (see~\cref{fig:landmarks}) and curves on the template model to guide the measurement algorithms. We take a total of 25 measurements on the body shapes, grouped into different classes of measurement types.

To measure the \emph{bust circumference}, \emph{hip circumference}, \emph{waist circumference} and \emph{leg circumference}, we use an edge loop defined on the template mesh as an initial guess.
We then compute a measurement plane that is parallel to the ground plane and minimizes the sum of squared distances to the vertices of this edge loop (see~\cref{fig:landmarks}). 
The convex hull of the intersection between the mesh and the measurement plane (a planar polyline) provides the required circumference. This mimics a tailor placing a tape measure around the body. It is possible for the arms to cross the measurement plane, so if multiple closed polylines are detected, the largest circumference is selected. 

Due to some amount of tangential drift in the template fitting process, the computed measurement plane may be slightly misplaced.
To remedy this, we shift the plane in its up-direction in the range [\SI{-2}{cm},\,\SI{+2}{cm}] in \SI{5}{mm} steps and measure the resulting convex hull length. The final measurement value is defined by the local maximum of these values for the \emph{bust circumference} and \emph{hip circumference} and the local minimum for the \emph{waist circumference}.
Subsequently, we recalculate the landmark positions for the \emph{waist side} (left and right), \emph{mid waist back}, \emph{bust} (left and right), \emph{bust side} (left and right), \emph{hip side} (left and right) and \emph{bum} (left and right), so that these landmarks are the closest point on the intersection polyline.
For the \emph{leg circumference} measurement, the measurement plane is shifted down until both legs are separated in this plane. We measure both leg circumferences and record the minimum value of both legs. 

For the \emph{back width}, the \emph{hip back width} and the \emph{waist back width} measurements, only the back-facing part of the circumference curve is relevant. We use the corresponding left and right landmarks on the body to define the coronal plane and measure the back-facing part of the convex hull described above. 

We use similar principles to compute the remaining measurements, as detailed in the supplemental material.

\subsection{Draping}
\label{sec:gen:drape}

We choose Warp~\cite{warp2022}, specifically its XPBD implementation, as our backbone simulator, since it is one of very few open-source and GPU-based simulators, providing the speed and scalability necessary for data generation. However, Warp includes only basic functionality, not specifically targeted for garment draping, so we have to solve multiple challenges to successfully simulate randomized garment designs at scale.\\

\setlength{\columnsep}{10pt}%
\setlength{\intextsep}{0pt}%
\begin{wrapfigure}{r}{0.3\linewidth}%
  \includegraphics[width=\linewidth]{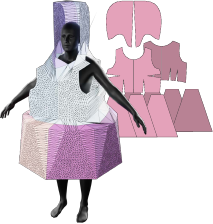}%
  \vspace{-0.2cm}
  \caption{Box mesh}
  \label{fig:boxmesh}%
\end{wrapfigure}%

\vspace{-2mm}
\noindent\textbf{Draping setup.\ }
Sampled sewing patterns are stored in a JSON-based format~\cite{Korosteleva2021} that describes a sewing pattern as a structure. To prepare it for simulation, we generate the corresponding mesh using a box mesh paradigm: we compute a single connected mesh from the pattern panels by connecting vertices of edge pairs participating in a stitch (\cref{fig:boxmesh}). The details of this process are provided in the supplemental material. 

\begin{figure}[t]
  \centering
  \subcaptionbox{Initial self-collisions\label{fig:sim_tips:self-col}}{
        \includegraphics[height=3.4cm]{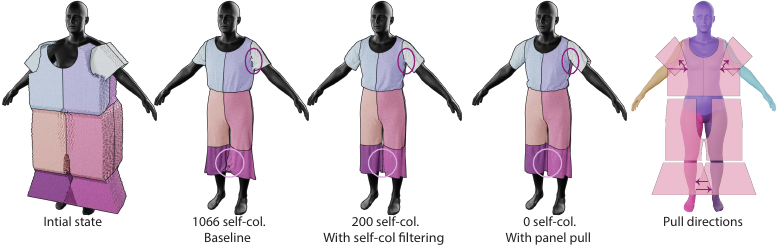}
    }\\
    \vspace{0.1cm}
    \subcaptionbox{Body collisions with pull\label{fig::sim_tips:body-drag}}{
        \includegraphics[height=3.2cm]{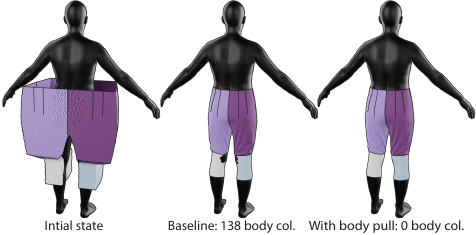}
    }
    \subcaptionbox{Body collisions filter\label{fig:sim_tips:body-filter}}{
        \includegraphics[height=3.2cm]{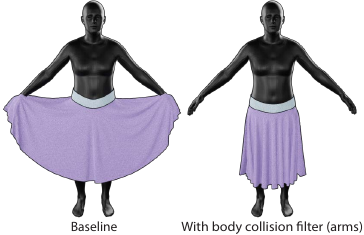}
    }\\
    \vspace{0.1cm}
    \subcaptionbox{Attachment: constraining garment waist vertices to be above body's waist level\label{fig::sim_tips:attachment}}{
        \includegraphics[height=3.2cm]{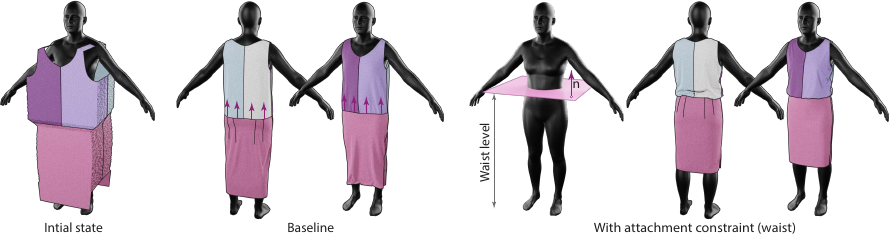}
    }
  \label{fig:sim_tips}
  \caption{Resolving simulation issues. \emph{Pull} refers to collision resolution constraints.}
\end{figure}

We set up the simulation of the box mesh by supplying the original mesh edge lengths as a reference, collapsing the edges elongated due to stitch vertices merging and effectively stitching the panels together. We also implement the point-triangle and edge-edge self-collision prevention algorithms for XPBD~\cite{Lewin2018ClothSC,cincotti2022}, which are not available in Warp. Similarly to Korosteleva and Lee~\cite{Korosteleva2021}, we allow the first few frames of the simulation to be gravity-free for the benefit of the stitching process. We also use similar conditions on the static equilibrium. If a small enough percentage of mesh vertices ($1.5\%$) change their location between the frames by a magnitude greater than a specified threshold (\SI{0.04}{\cm}), the mesh is considered static. With the addition of material properties selection, this constitutes the simplest draping setup. However, due to the diversity of the pattern designs and the heuristic initial  placement of the pattern panels, many examples would be draped incorrectly (\cref{fig:sim_tips}). Below we describe our approach to more robust, automatic draping.\\

\vspace{-2mm}
\noindent\textbf{Resolving initial collisions.\ }
One of the major challenges in draping is the collision-free initial state requirement. Collision handling methods are typically designed to prevent new collisions from occurring, but cannot resolve collisions that already exist in the initial state. In our scenario, the designs are varied automatically, and the panel placement is only approximately assigned, so both self- and body-collisions are inevitable. Surprisingly little work addresses this issue when applied to cloth simulation. {Existing approaches either tackle special cases irrelevant to us~\cite{Baraff2003UntanglingCloth,Buffet2019ImplicitClothing} or use optimization-based solutions~\cite{Volino2006ResolvingMinimization,Ye2017ADetection} that complicate the system and significantly slow the draping}. We propose a selection of methods that efficiently handle different types of initial collisions using XPBD-friendly techniques. 

We note that the standard self-collision resolution methods aim to preserve the existing relative state of the garment mesh, counteracting attempts to resolve self-collisions by other means. We propose to perform global collision detection by finding the mesh edges that intersect non-adjacent faces using ray-intersect checks and then excluding these mesh edges and their adjacent vertices from the self-collision prevention mechanism, effectively allowing these elements to pass through the garment mesh freely. This approach already allows for successful collision-free draping of box meshes akin to spring-based pattern stitching, but some artifacts remain, as demonstrated in~\cref{fig:sim_tips:self-col}. 

For the initial self-collisions, the hardest problem is determining the direction of pulling on the mesh vertices to resolve the collisions. We notice that most of the overlaps between the panels of the same pattern occur in the areas between body parts, where little space is available: the trunk and the arms, or between the legs. Our idea is to introduce an additional constraint that pulls apart the intersecting geometry towards the body parts they semantically correspond to, thereby eventually resolving the collisions (\cref{fig:sim_tips:self-col}). This constraint is applied to vertices of edges that intersect the garment mesh faces belonging to a different body part compared to the edge in question. 

To determine panel-body part correspondence, we utilize the fact that we have access to the pattern modeling process from GarmentCode~\cite{Korosteleva2023GarmentCode:Patterns} and that all the body meshes share the same topology. First, we define a simple segmentation mask on the body indicating each limb (left arm, right arm, left leg, right leg) and trunk, as shown in~\cref{fig:sim_tips:self-col} (right). In addition, we define combinations of limbs (both arms, both legs) for convenience. We then match every panel and its corresponding mesh vertices to one of the indicated body parts by calculating the body part closest to most of the vertices of that panel or using the panel label if provided. 
Note that only the pant panels are matched with the right and left legs separately, while skirts are matched to both legs as a single unit. 

Mesh-body segmentation labels allow us to define collision filters: it is unlikely that the lower body garments collide with the arms, so to prevent wide skirts from being stuck in the arm area, as in the example in~\cref{fig:sim_tips:body-filter}, we ignore the collisions between arms on the body mesh and the parts of the garments mesh associated with the legs, allowing skirts to fall through the arms freely. We use the same mechanism to discard collisions with internal geometry pieces of the body (mouth and eyeballs).

\setlength{\columnsep}{10pt}%
\setlength{\intextsep}{0pt}%
\begin{wrapfigure}{r}{0.3\linewidth}%
  \includegraphics[width=\linewidth]{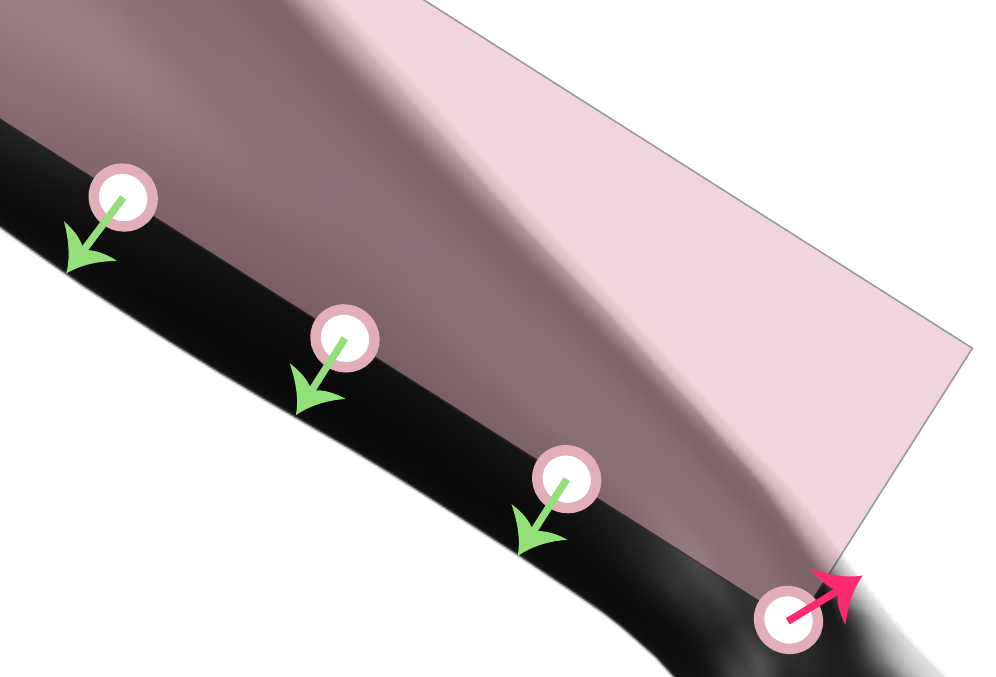}%
  \vspace{-0.2cm}
  \caption{Body-cloth constraint. Red arrow indicates an erroneous match.}
  \label{fig:body_drag_close}%
\end{wrapfigure}%

Finally, we apply a constraint that pulls the vertices of the garment mesh found inside the body mesh towards the closest point at the surface of the body, see~\cref{fig:body_drag_close}. Granted, this simple approach may match the garment point to the body surface incorrectly resulting in worsening the body-garment collisions (\cref{fig:sim_fails:arms_stuck}), but we found it helpful on average in our design drapes, see~\cref{sec:eval:drag}. We assume that thanks to reasonable placement initialization, most of the garment vertices erroneously placed inside the body are not too far from the correct side of the body.\\

\vspace{-2mm}
\noindent\textbf{Attachment constraints.\ }
Errors in panel placement not only cause collision issues, but sometimes lead to incorrect placement of garment elements on the body in the final simulation. We employ attachment constraints on the lower-body garments (pants, skirts) to force their placement on the waist area, to pull strapless tops up towards the underarm level, and to pull neckline vertices closer to the neck, which effectively places the shoulder area of the garment at the correct location. The attachment constraints are released after a set amount of frames to allow position adjustment towards the natural location (e.g., low-rise bottoms should be placed lower than the waist). 

To simplify attachment constraint specification, we propose what we call a \emph{half-space attachment constraint}, where instead of specifying a desired location for each vertex participating in the constraint, we specify the target half-space, defined by a plane and its normal direction. With this formulation, the waist constraint can be expressed as simply having the 
vertical coordinate of the affected vertices to be higher than the waistline of the body shape, which is agnostic to the mesh topology of the body model, see~\cref{fig::sim_tips:attachment}. We use the known body measurements information to define the half-spaces for the attachment constraints: waist level for lower garments, underarm level for strapless tops, both with the vertical normal direction, and half of the neck width to the left and to the right of the head for collar vertices, with the normal pointing towards the center of the body along the horizontal axis.\\

\begin{figure}[t]  
  \centering
  \includegraphics[width=0.8\linewidth]{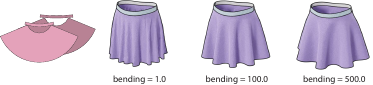}
  \caption{{Simulation materials used in GarmentCodeData.}}
  \label{fig:fabric_options}
\end{figure}

\vspace{-2mm}
\noindent\textbf{Material properties.\ } 
We fix the stiffness across all samples to have minimal stretching. To add variation in materials, we use different levels of fabric softness by varying the bending properties (\cref{fig:fabric_options}). The values are chosen manually to cover typical behavior of fabrics in daily wear. Materials are assigned to samples randomly, each covering approximately a third of the data points.\\

\vspace{-2mm}
\noindent\textbf{Quality control.\ }
Due to the significant diversity in design samples, some may not drape correctly and hence need to be excluded from the dataset to avoid confusing trained models with unrealistic data points. We limit the simulation of each sample by time (5 minutes) and number of frames (2400). If the static equilibrium is not achieved, e.g., a lower garment has slid off and keeps falling, the example is marked as a fail. We also filter out examples with a high number of cloth-body and self-collisions, counted as the number of garment mesh edges intersecting the body mesh or the garment mesh itself, respectively. The success rate of simulation, accounting for failures of mesh generation and quality control, is about 72\%. We analyze the failure cases in~\cref{sec:eval:sim_fails}.\\

\vspace{-2mm}
\noindent\textbf{Generation speed.\ }
We create the dataset in batches of 5,000 design samples each. The batches are independent of each other and can be processed in parallel. Sewing patterns that fail the quality checks are discarded, and the generation continues until the desired number of correct samples is reached. It takes about 7 hours to obtain a set of 5,000 designs, fitted for both random body shapes and the neutral body, making it 10,000 sewing patterns in total that passed the quality checks when running on one core of AMD EPYC 9654 CPU.

Each batch is then processed in the draping step, running a simulation on the neutral body and random body shapes in parallel, 5,000 drapes for each. {The body mesh consists of 47.5k triangles and each garment contains about 30k triangles on average.} When running on NVIDIA GeForce RTX 3090, each sample takes 30 seconds to simulate on average, amounting to about 44 hours for 5,000 drapes, with about 2 hours overhead for renders and other processing, resulting in approximately 3,200 final designs that pass the quality check per batch.  

\section{Analysis}

\subsection{Body measurements quality}

In order to evaluate the accuracy of our automatic measurements, we evaluate those measurements that have a counterpart in the CAESAR data set. These measurements have been manually taken by tailors and are assumed to be the ground truth in our evaluation. 
The measurements are on average overestimated by \SI{2.81}{cm} for the \emph{bust circumference}, \SI{1.99}{cm} for the \emph{hip circumference} and \SI{3.49}{cm} for the \emph{leg circumference}.
High errors arise due to possible tangential drift (bust measurement) and when the scanned individuals are wearing wide, non-skin-tight pants ({leg circumference}).
Overestimations in our measurements may also arise due to the missing compression of soft tissue when measuring a static mesh. The \emph{shoulder width} is underestimated on average by \SI{4.57}{cm} our method, as we measure on top of the shoulder ball joint, {where the garment shoulder line is typically located}. The ground truth measurement is taken on the left and right sides of the deltoid muscles. 
The overestimation of the circumference measurements does not have a negative impact on the fitting of the garments in our experiments.\\

\begin{figure}[t]  
  \centering
  \includegraphics[width=0.85\linewidth]{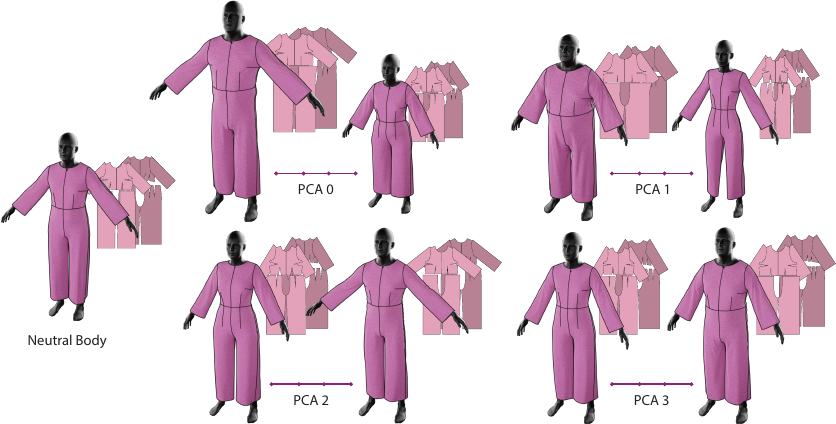}
  \caption{Design transfer across extreme PCA values based on automatic measurements.}
  \label{fig:body_fits}
\end{figure}

\vspace{-2mm}
\noindent\textbf{Garment fitting across body shapes.\ }
To showcase the success of our automatic measurement-taking algorithms, we retarget a fitted jumpsuit pattern to a selection of bodies corresponding to the most extreme PCA values present in the body shape samples (\cref{fig:body_fits}). The fitted jumpsuit sewing pattern covers all of the body curves on the body trunk and hence is a suitable design to evaluate the quality of tight fit. In all the examples  used, the garments follow the body shape smoothly without excessive wrinkling or balance line distortions, indicating correct fit. The quality of retargeting is almost surprisingly good for heuristic-based body measurements and parametric sewing patterns. 

\subsection{Simulation failure cases}
\label{sec:eval:sim_fails}

\begin{figure}[t]  
  \centering
  \subcaptionbox{\label{fig:sim_fails:shoulder_slide}}{
    \includegraphics[trim={0 0 0 4.5cm},clip,height=2.7cm]{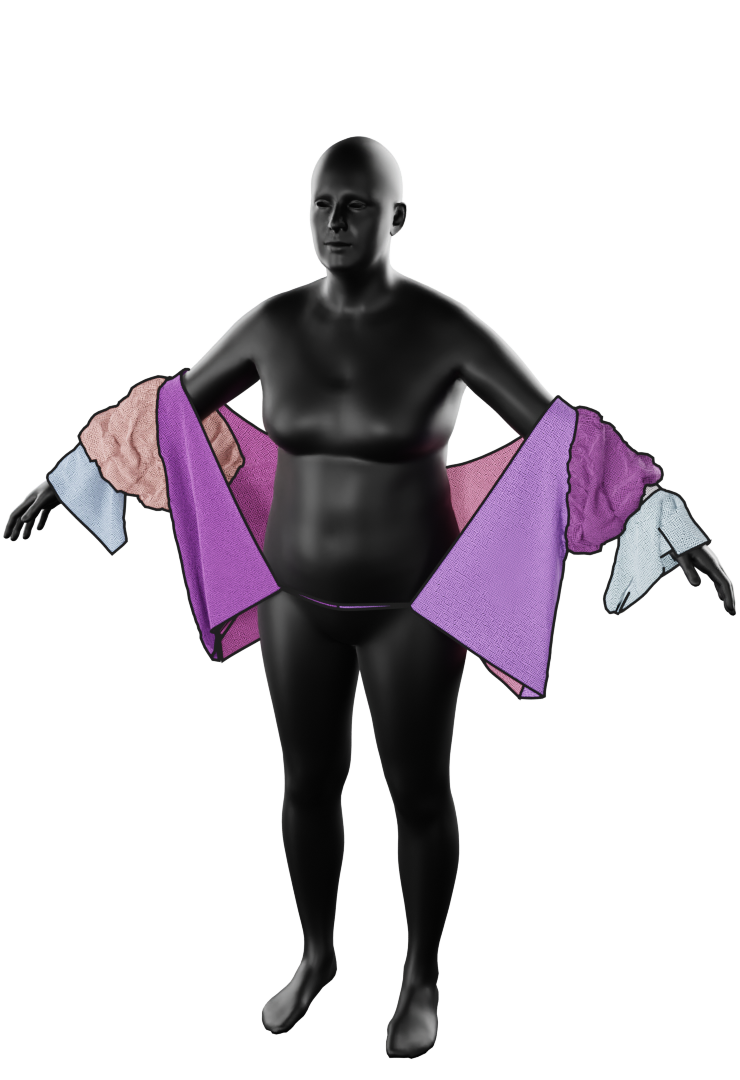}
    }
    \hspace{-0.45cm}
    \subcaptionbox{\label{fig:sim_fails:full_slide}}{
    \includegraphics[trim={0 0 0 4.5cm},clip,height=2.7cm]{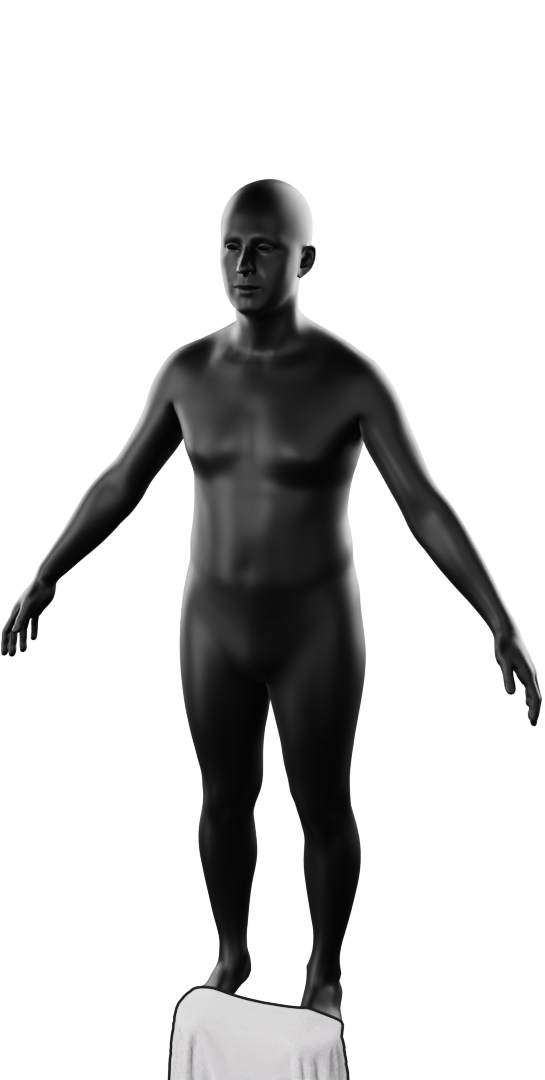} %
    }
    \hspace{-0.45cm}
    \subcaptionbox{\label{fig:sim_fails:strapless_slide}}{
    \includegraphics[trim={0 0 0 4.5cm},clip,height=2.7cm]{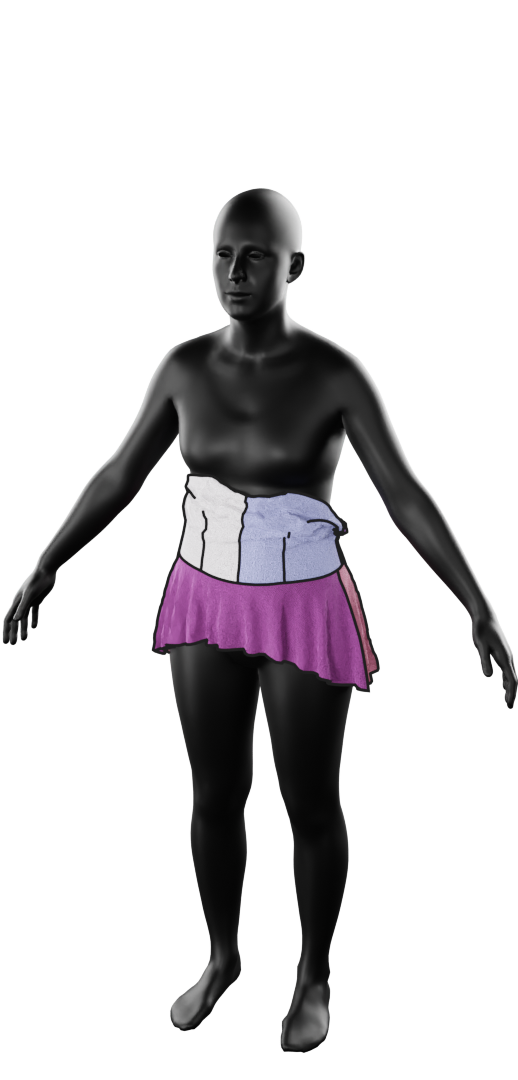}
    }
    \hspace{-0.55cm}
    \subcaptionbox{\label{fig:sim_fails:overcollisions}}{
    \includegraphics[trim={0 0 0 4.5cm},clip,height=2.7cm]{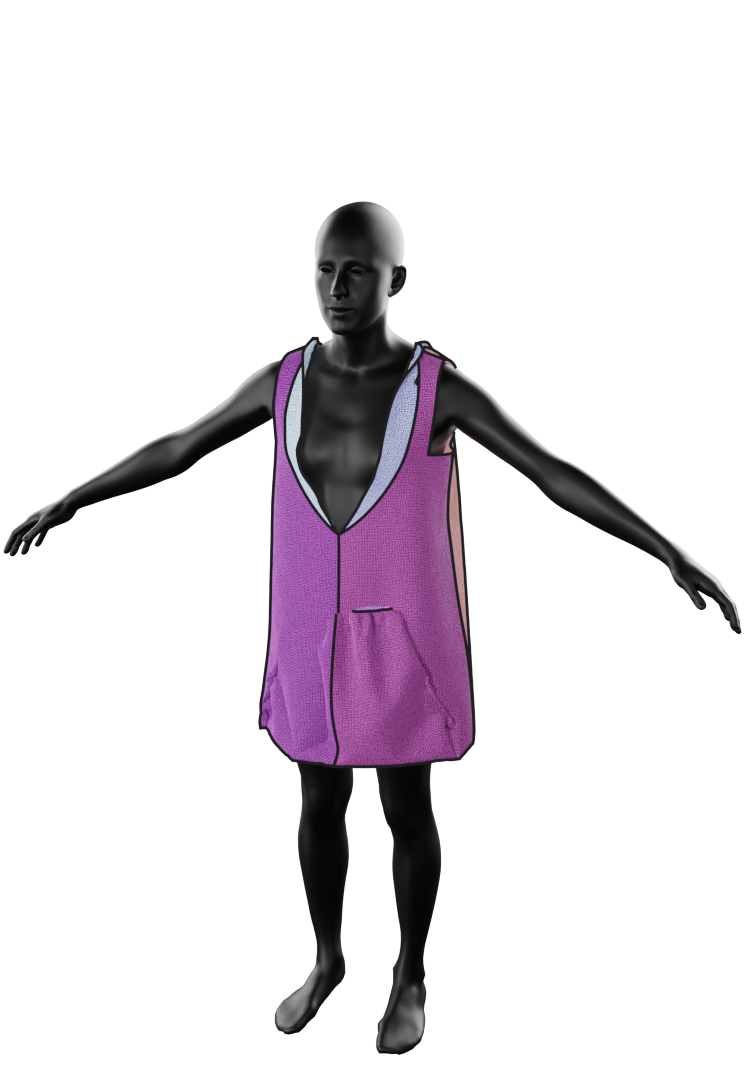}
    }
    \hspace{-0.45cm}
    \subcaptionbox{\label{fig:sim_fails:cuffs_out}}{
    \includegraphics[trim={0 0 0 4.5cm},clip,height=2.7cm]{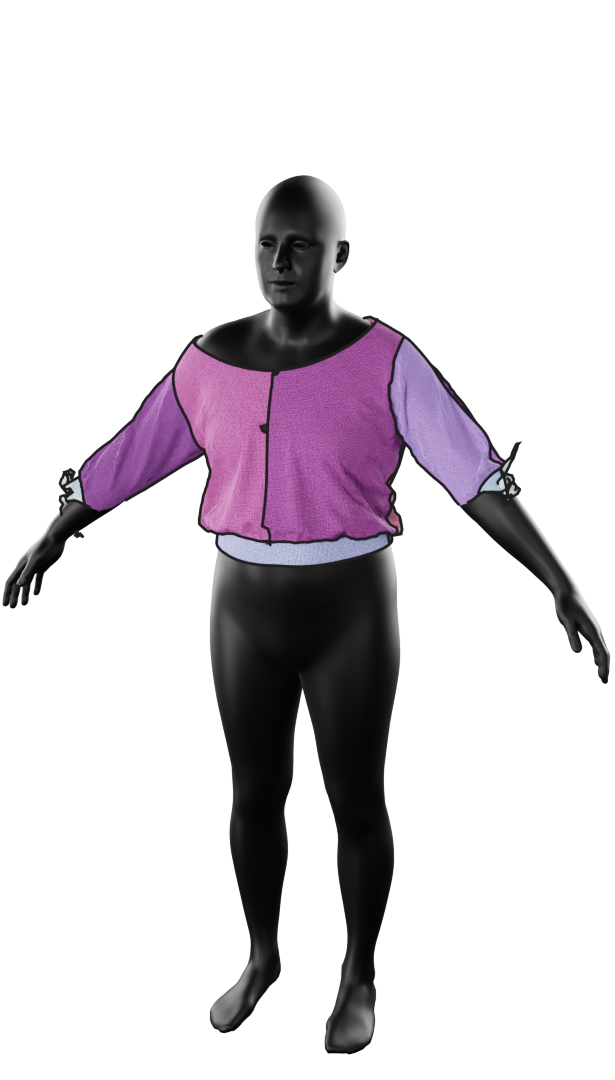}
    }
    \hspace{-0.45cm}
    \subcaptionbox{\label{fig:sim_fails:feet_stuck}}{
    \includegraphics[trim={0 0 0 4.5cm},clip,height=2.7cm]{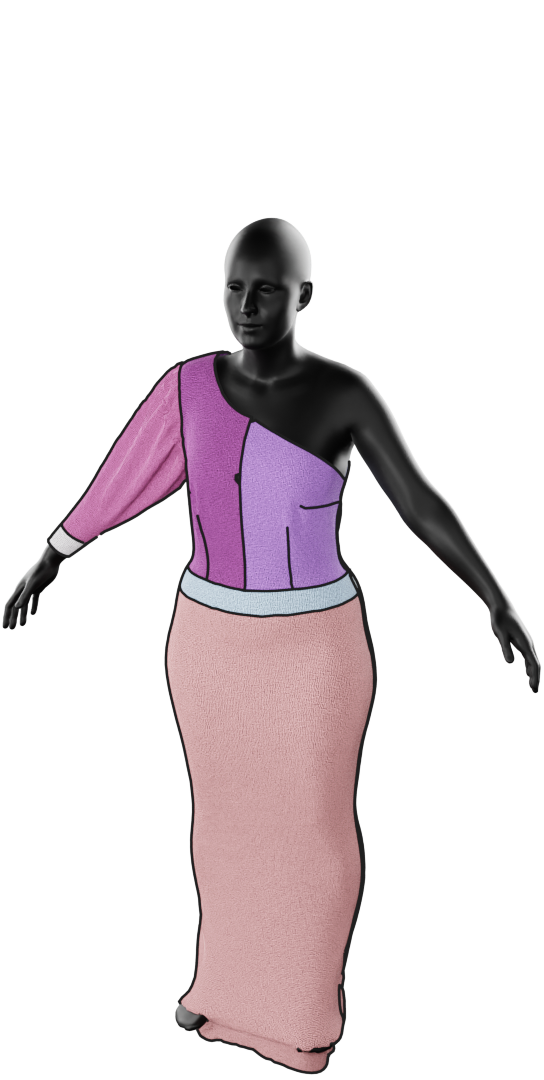}
    }
    \hspace{-0.45cm}
    \subcaptionbox{\label{fig:sim_fails:arms_stuck}}{
        \includegraphics[trim={0 0 0 4.5cm},clip,height=2.7cm]{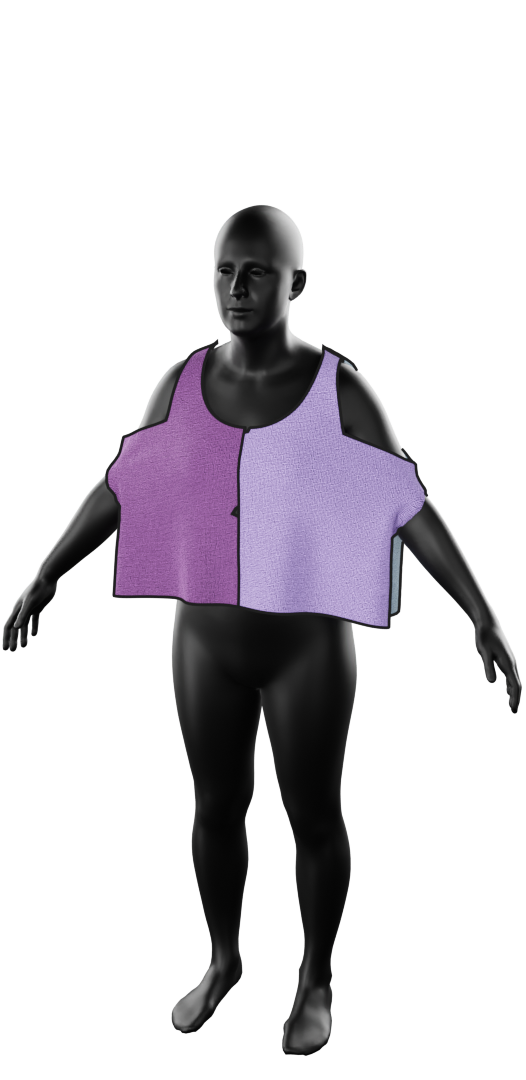}
    }
  \caption{Simulation failure cases.}
  \label{fig:sim_fails}
\end{figure}  

\cref{fig:sim_fails} showcases typical failures of the simulation process.  \cref{fig:sim_fails:shoulder_slide,fig:sim_fails:full_slide,fig:sim_fails:strapless_slide} show sliding errors, occurring when the garment is bulky and heavy, while the sewing pattern design does not allow for proper grip on the given body shape. For example, the low-rise circle skirt in \cref{fig:sim_fails:full_slide} falls down due to low body curvature and material stretch. Such designs are likely to be impractical in real life context, so we consider these failures as legitimate. \cref{fig:sim_fails:overcollisions} demonstrates heavy self-intersections of a skirt element with a long upper-body piece, failing to preserve a non-colliding state during the application of attachment constraints. While this example reveals weaknesses in collision resolution, one could argue that this design is unrealistic to begin with, since such a top would normally be disconnected from the bottom and flow over it, or the skirt would not be fitted to sit on the waist, but rather hang at the bottom in a dropped waistline style.  \cref{fig:sim_fails:cuffs_out,fig:sim_fails:feet_stuck} are dressing failures, where our method failed to resolve body-cloth misalignment, leaving cuffs sticking out (\cref{fig:sim_fails:cuffs_out}) or skirt caught in the feet (\cref{fig:sim_fails:feet_stuck}). \cref{fig:sim_fails:arms_stuck} is a similar case and a likely result of an erroneously applied body-cloth collision constraint, when the wide shirt trunk has initial intersections with the arms after draping, and the internal parts are incorrectly pulled towards the top of the arm. These failure cases demonstrate limitations of hand-crafted design spaces, as well as a need for better body-cloth collision resolution due to initial pattern placement.

\subsection{Simulation pipeline ablation study}
\label{sec:eval:drag}
\begin{table}[bt]
\centering
\caption{Evaluation of collision resolution constraints (\emph{pull}) for drapes on neutral and random body shapes. $k^\textit{th}$ stands for $k^\textit{th}$ percentile.} 
\label{tab:drag_collisions}
\begin{tabular}{lcccccc}
\toprule
               & \multicolumn{3}{c}{\#Body collisions} & \multicolumn{3}{c}{\#Self-collisions} \\
               & Median       & $80^\textit{th}$ & $95^\textit{th}$      & Median       & $80^\textit{th}$       & $95^\textit{th}$      \\ \midrule
neutral w/o pull   &  0           &  0        &   82.8   &  10          &  162      & 747      \\
neutral with pull  &  0           &  0        &   52.2   &  0           &  104      & 423       \\ \hline
random w/o pull  &  0           &  24       &  141     &  6           &  200      &  535     \\
random with pull &  0           &  10       &  71.2    &  0           &  45       &  346     \\ \bottomrule
\end{tabular}
\end{table}

As~\cref{fig:sim_tips} demonstrates, attachment constraints and body collision filters show qualitative improvements for cases they are designed for, hence the need for their use is self-evident. However, the proposed constraint for body-cloth collision resolution has known failure cases when vertices are matched to the wrong side of the body. Hence, we set up an experiment where we run a simulation on the same dataset of 100 samples with and without additional collision constraints, reporting the statistics for body-cloth and, for completeness, cloth-cloth collisions across the simulated samples (including failure cases). \cref{tab:drag_collisions} clearly shows the decrease in the number of collisions when constraints are enabled, showing the usefulness of the constraints on average.

\section{Discussion and limitations}

We presented a generation pipeline for sampling 3D made-to-measure garments with ground-truth sewing patterns, as well as GarmentCodeData -- a dataset of 115,000 3D garments of different designs fitted to a variety of body shapes. Although we aimed to create a large-scale dataset suitable for a sewing pattern reconstruction task, we envision other potential applications. The proposed probability-based design space control allows for its easy adjustment and creation of specialized design sets, for example, specific garment types (workwear, uniforms, etc.). The availability of various designs fits to a neutral body shape in the dataset opens up applications for design retargeting. Using the pipeline for neural cloth simulation research could also be an interesting direction to explore.\\ 

\vspace{-2mm}
\noindent\textbf{Limitations.\ }
Our work improves on some of the limitations of existing works, but there is still ample room for creating truly comprehensive datasets. First, we inherit some limitations of the GarmentCode~\cite{Korosteleva2023GarmentCode:Patterns} patterns, addressing which would require additional effort put into the draping pipeline: lack of fabric layers, sharp pleats, hardwear (e.g., zippers, buttons or hooks), the ability to represent full outfits. Our take on fabric material distribution is rather minimalistic, and further effort is needed to model materials according to real-world fabric properties. As discussed in~\cref{sec:eval:sim_fails}, some of the simulation errors we encounter are due to issues with initial body-pattern alignment and methods for body-cloth collision resolution, which require further research. 
As for body measurements, our approach may result in inaccurate measurements due to errors in surface fit, especially on extreme body shapes. Such errors may lead to the topology-based initial guess being far off, and the correct horizontal plane may not be found within the given optimization bounds. \\

\vspace{-2mm}
\noindent\textbf{Ethical considerations.\ }
While CAESAR~\cite{robinette2002caesar} is one of the best collections of 3D human body shape data, which is of good quality and responsibly collected, it mainly covers the body shapes of healthy European and North American adults, hence the statistical body model is biased towards these body shapes and would have trouble representing geographic variations in body sizes, people with disabilities, teens and children, transferring this bias into the 3D garment samples. Due to the price point on the full dataset, we only use the European subset of CAESAR, which further worsens the problem. The unavailability of open, unbiased 3D human datasets is a critical issue that propagates biases into many downstream applications. 

Our work required a sizable amount of computational power to generate the data.
However, the server infrastructure we use is carbon neutral, and is running fully on renewable energy~\cite{cscs_sustainability}, which minimizes the climate impact of this work.

\subsubsection{Acknowledgments.}

This work was supported by the European Research Council (ERC) under the European Union’s Horizon 2020 Research and Innovation Programme (ERC Consolidator Grant, agreement No.\ 101003104, MYCLOTH). Stephan Wenninger has been funded by ``Stiftung Innovation in der Hochschullehre'' through the project ``Hybrid Learning Center'' (FBM2020-EA-690-01130). We thank  Ami Beuret for his help in refactoring GarmentCode codebase. We are grateful to Jana Schuricht for her professional consultations on patternmaking and to members of IGL and GGG for their continuous support and cheer throughout this project.

%
\bibliographystyle{splncs04}
\bibliography{bibliography,appendix,mend_copy}

\clearpage
\appendix

\renewcommand\thefigure{\thesection.\arabic{figure}}
\setcounter{figure}{0}

\section*{GarmentCodeData: Supplemental Material}

\section{Body shape sampling}

Principal component analysis (PCA) on the registered meshes (\cref{sec:gen:body}) gives us a generative model to sample various body shapes. 
Let $V$ be the number of vertices in the template model, $\pcamat \in \R^{3V \times k}$ contain the principal components scaled by the square root of their respective eigenvalues, and $\pcamean \in \R^{3V}$ correspond to the mean vertex positions of the dataset. 
New shapes can then be sampled by generating a parameter vector $\pcaparams \in \R^{k}$ and reconstructing new vertex positions $\mat{X} = \pcamat\pcaparams + \pcamean$. 

We sample 5000 shapes by independently sampling each PCA parameter $w_i$ following a normal distribution with $\mu = 0$ and $\sigma = 0.7$, such that the probability of generating $w_i \in (-2.4, 2.4)$ is $\SI{99.94}{\percent}$. 
After generating 5000 shapes, we export each body model in two different poses. 
We vary the arm inclination of the model by drawing a random angle from a uniform distribution over $[\SI{-15}{\degree}, \SI{15}{\degree}]$ and altering the standard A-pose of the model accordingly. 
Additionally, we export the same shape in a pose where the feet are slightly more apart from each other (see~\cref{fig:teaser}).
The body measurements described in~\cref{sec:gen:body} are all taken in the first of the two poses.

\section{Taking measurements}

Here, we provide the full description of our measurement-taking algorithms, following the principles outlined in~\cref{sec:gen:body}. We include the ones already described in~\cref{sec:gen:body} for completeness. 

To measure the \emph{bust circumference}, \emph{hip circumference}, \emph{waist circumference} and \emph{leg circumference}, we use an edge loop defined on the template mesh as an initial guess.
We then compute a measurement plane that is parallel to the ground plane and minimizes the sum of squared distances to the vertices of this edge loop (see~\cref{fig:landmarks}). 
The convex hull of the intersection between the mesh and the measurement plane (a planar polyline) provides the required circumference. This mimics a tailor placing a tape measure around the body. It is possible for the arms to cross the measurement plane, so if multiple closed polylines are detected, the largest circumference is selected. 

Due to some amount of tangential drift in the template fitting process, the computed measurement plane may be slightly misplaced.
To remedy this, we shift the plane in its up-direction in the range [\SI{-2}{cm},\,\SI{+2}{cm}] in \SI{5}{mm} steps and measure the resulting convex hull length. The final measurement value is defined by the local maximum of these values for the \emph{bust circumference} and \emph{hip circumference} and the local minimum for the \emph{waist circumference}.
Subsequently, we recalculate the landmark positions for the \emph{waist side} (left and right), \emph{mid waist back}, \emph{bust} (left and right), \emph{bust side} (left and right), \emph{hip side} (left and right) and \emph{bum} (left and right), so that these landmarks are the closest point on the intersection polyline.
For the \emph{leg circumference} measurement, the measurement plane is shifted down until both legs are separated in this plane. We measure both leg circumferences and record the minimum value of both legs. 

For the \emph{back width}, the \emph{hip back width} and the \emph{waist back width} measurements, only the back-facing part of the circumference curve is relevant. We use the corresponding left and right landmarks on the body to define the coronal plane and measure the back-facing part of the convex hull described above. 

To take the \emph{waist over bust line} measurement, we use a parasagittal plane with the least squares distance to a predefined edge loop on the template mesh. We calculate the polyline of the cut plane and the given mesh and consider a part of this polyline for the final measurement. The intersection plane computed when measuring the \emph{waist circumference} defines the lowest point.
The upper boundary is defined by the frontal plane with a minimum distance to the \emph{shoulder} landmarks. We compute the \emph{bust line} similarly, starting at the intersection plane from the \emph{bust circumference} measurement (see~\cref{fig:landmarks}).

We use the Euclidean edge loop length to measure the neck width and wrist circumference. 
For the \emph{armscye depth} measurement and the Euclidean distance between the \emph{bum points}, the \emph{bust points} and the \emph{shoulder width}, we compute the respective distances between the two landmarks for each measurement. 
The \emph{crotch hip difference}, \emph{hip line} and the \emph{vertical bust line} are measured as distances between two landmarks along the upward axis. 
For the \emph{height} measurement, the axis-aligned bounding box of the mesh is determined. The final value is the length of the bounding box in the upright dimension.
We measure the \emph{head length} as the distance between the neck landmark and the highest point of the head along the upward axis.
We calculate the \emph{arm length} as the geodesics distance between the \emph{shoulder} and \emph{wrist} landmarks.

We measure the \emph{arm inclination} and \emph{hip inclination} angles as the opening angle between the median plane of the body and a vector given by the two landmarks. 
Finally, we calculate the \emph{shoulder inclination} as the angle between the horizontal plane and a vector given by the \emph{neck} and \emph{shoulder} landmarks.

\section{Mesh generation}
\begin{figure}[ht]  
  \centering
  \includegraphics[width=1.\linewidth]{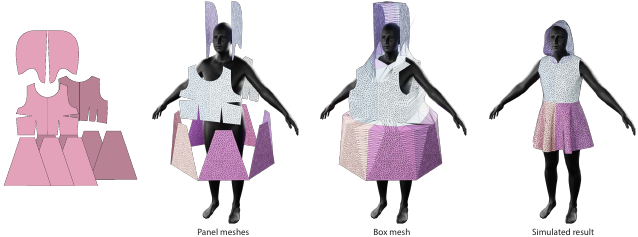}
  \caption{Mesh generation process}
  \label{fig:mesh_gen}
\end{figure}

Sewing patterns coming out of GarmentCode~\cite{Korosteleva2023GarmentCode:Patterns} are stored in the specialized JSON-based format~\cite{Korosteleva2021} that represents a sewing pattern as a structure. The pattern is described as a collection of panels, which in turn are specified through the shape of their outline, a looped sequence of directed curved edges connecting vertices on the panel border, with additional specification of the panel 6D placement that locates it in 3D. The specification also contains stitching information as a list of edge pairs connected by a stitch. This representation is unsuitable for simulation, so the first step of the draping process is generating a mesh corresponding to the input sewing pattern. We follow the path of generating a box mesh: generating a fully connected mesh for the whole sewing pattern by directly merging the vertices of the edge pairs connected by a stitch (\cref{fig:mesh_gen}). First, we generate vertices on every panel edge by placing them on edge curves according to a chosen mesh resolution (here we aim for a \SI{1}{\cm} average mesh edge length). Panel edges connected by a stitch may have different lengths (for a gathering/ruching effect), in which case we match the number of vertices on either side by using the maximum number of vertices suggested by the resolution among the two edges, and using this number for both, spacing them equally along the lengths. Once the edge vertices are fixed, we generate internal panel vertices using constrained Delaunay triangulation of CGAL~\cite{cgal:eb-23b}. The vertices are placed in 3D using the panel rotation and translation information from the JSON specification. Triangle normals are set to match the panel normal defined by the counter-clockwise orientation of the edge loop in the specification. Stitches are then processed by merging the matching vertices on either side of a stitch and placing them in the middle between the original locations while keeping the information on the original length of the edges to correctly reduce the stitching edges' lengths during the simulation step.

\end{document}